# DETECTING FAKE JOB POSTINGS USING BIDIRECTIONAL LSTM

**Aravind Sasidharan Pillai[*1]**

[*1]The University Of Illinois, Urbana-Champaign, IL, USA.

## ABSTRACT

Fake job postings have become prevalent in the online job market, posing significant challenges to job seekers and employers. Despite the growing need to address this problem, there is limited research that leverages deep learning techniques for the detection of fraudulent job advertisements. This study aims to fill the gap by employing a Bidirectional Long Short-Term Memory (Bi-LSTM) model to identify fake job advertisements. Our approach considers both numeric and text features, effectively capturing the underlying patterns and relationships within the data. The proposed model demonstrates a superior performance, achieving a 0.91 ROC AUC score and a 98.71% accuracy rate, indicating its potential for practical applications in the online job market. The findings of this research contribute to the development of robust, automated tools that can help combat the proliferation of fake job postings and improve the overall integrity of the job search process. Moreover, we discuss challenges, future research directions, and ethical considerations related to our approach, aiming to inspire further exploration and development of practical solutions to combat online job fraud.

## I. INTRODUCTION

The rapid growth of the internet has transformed the way job seekers and employers interact, with online job portals becoming a vital resource for millions of people worldwide. However, while these platforms offer numerous benefits, they have also raised a significant problem: the proliferation of fake job postings. Fraudulent job advertisements waste job seekers' time and resources, posing severe risks such as identity theft and financial loss. Therefore, effective methods are needed to detect and mitigate the impact of fake job postings.

Machine learning and NLP techniques have shown great promise in detecting deceptive content across various domains, including spam email detection, fake news identification, and sentiment analysis. In this context, Recurrent Neural Networks (RNNs) and their variants, such as Long Short-Term Memory (LSTM) networks, have emerged as powerful tools for processing sequential data and capturing the temporal patterns in text. Furthermore, Bidirectional LSTM (Bi-LSTM) networks have demonstrated remarkable performance in various NLP tasks, as they can learn and process contextual information from past and future time steps.

This paper proposes a novel approach to detecting fake job postings using Bidirectional LSTM networks. We hypothesize that the ability of Bi-LSTM to capture the complex structure of textual data effectively can be harnessed to distinguish between genuine and fraudulent job advertisements. We present a comprehensive methodology, including text preprocessing, word embedding, and model training, and evaluate our proposed model on various datasets. Finally, through a series of experiments, we demonstrate the efficacy of our approach and compare its performance with other state-of-the-art techniques, ultimately showcasing the potential of Bidirectional LSTM in addressing the growing issue of fake job postings.

## II. METHODOLOGY

This section presents the methodology for detecting fake job postings using Bidirectional LSTM networks. Our approach comprises several key stages, including literature reviews, data analysis and preprocessing, and word embedding.

**2.1 Literature Review**

Detecting fake job postings is closely related to the broader field of deceptive content detection. In this section, we review several critical studies and techniques employed in detecting misleading content, including spam, fake news, and online reviews, as well as previous attempts to identify fake job postings.

Spam emails often contain fraudulent content and deceptive language. Researchers have developed various machine-learning algorithms to detect and filter spam emails. For example, Sahami et al. (1998)[1] proposed the use of Naive Bayes classifiers, while Drucker et al. (1999)[2] applied Support Vector Machines (SVM) for spam email classification. Both methods demonstrated promising results in spam detection. Detecting fake news is another application of deceptive content detection. Zhou and Zafarani (2018)[3] proposed a hybrid approach that combined linguistic features with user behavior analysis, achieving better performance than traditional machine learning methods. Online reviews can be manipulated, resulting in fake reviews that deceive consumers. Researchers have employed various NLP techniques to identify deceptive online reviews. Ott et al. (2011)[4] used a combination of n-grams and SVM to detect fake reviews on hotel websites.

In contrast, Li et al. (2020)[5] leveraged linguistic cues and sentiment analysis to detect deceptive reviews on e-commerce platforms. Previous research on seeing fake job postings has primarily focused on using traditional machine learning algorithms. For example, studies by Chen et al. (2018)[6] and Gupta et al. (2019) [7]focused on extracting relevant features from job postings, including textual, categorical, and numeric data. Zhang et al. (2021) [8] applied deep learning models to detect fake job postings. Although this study showed improved performance compared to traditional methods, they still needed to fully utilize the potential of deep learning in capturing sequential patterns.

A thorough literature review was performed, and several other classification studies [9]–[11] were analyzed to construct the baseline models.

### 2.2 Description of Data

For this study, we utilize the "Real / Fake Job Posting Prediction" dataset available on Kaggle. This dataset comprises 17,880 job postings, with each entry containing a mix of structured and unstructured data. The dataset is labeled, with 16,244 genuine job postings and 1,636 fake job postings, which makes it suitable for supervised learning tasks. Data characteristics are given in table 1.

**Table 1.** Data characteristics.

| Sl | Feature | Description |
|---|---|---|
| 1 | job_id | A unique identifier for each job posting. |
| 2 | location | The geographical location of the job. |
| 3 | department | The department or organizational unit of the job belongs |
| 4 | salary_range | The salary range for the job. |
| 5 | company_profile | A brief description of the company. |
| 6 | description | The detailed job description. |
| 7 | requirements | A list of required skills or qualifications for the job. |
| 8 | benefits | The benefits offered by the company. |
| 9 | telecommuting | A binary variable indicates whether the job allows telecommuting. |
| 10 | has_company_logo | A binary variable indicates a company logo's presence in the job posting. |
| 11 | has_questions | A binary variable indicates whether the job posting includes screening questions. |
| 12 | employment_type | The type of employment |
| 13 | required_experience | The required education level for the job. |
| 14 | industry | The industry to the job belongs. |
| 15 | function | The job function or role. |
| 16 | fraudulent | Target variable. A binary label indicating whether the job posting is genuine (0) or fake (1). |

### 2.3 Exploratory Data Analysis

Data exploration is integral to model building as it provides insight into the data and changes or modifications needed before designing the model. The distribution of binary variables is given in figure 1.

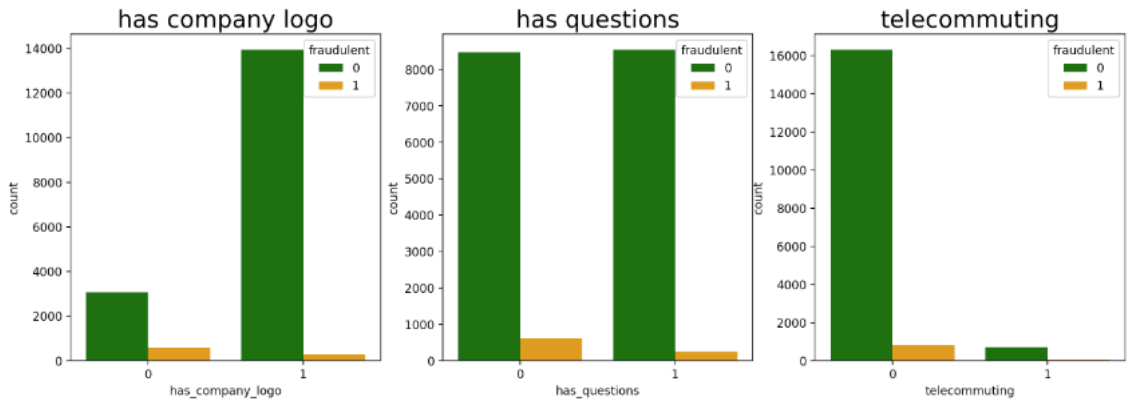

**Figure 1.** Binary variables distribution.

We did extensive textual data analysis using the CountVectorizer utility. CountVectorizer is a Natural Language Processing (NLP) technique used to convert a collection of text documents into a matrix of token counts. In the context of analyzing fake job postings, it can help identify common words, phrases, or patterns that may be associated with fraudulent listings. The word distributions for job titles are given in figure 2. Manager, developer, and engineer are the most frequently used words in the job title.

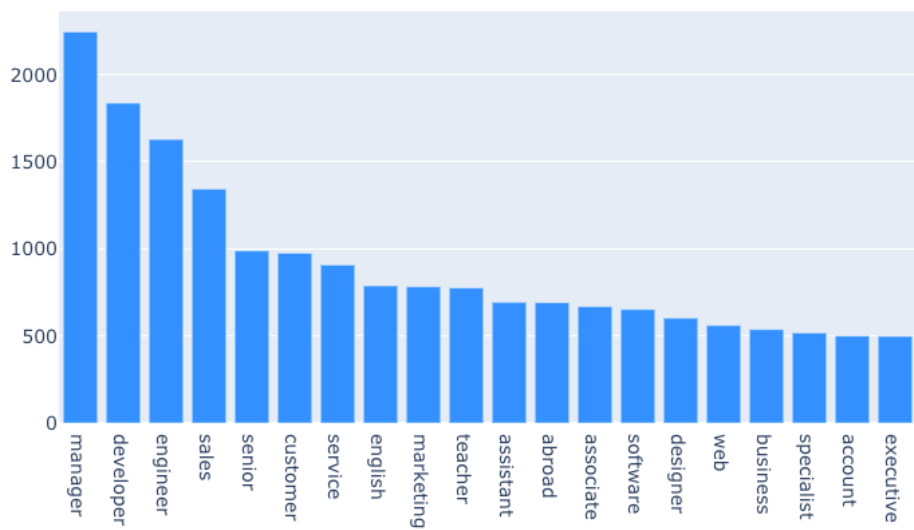

**Figure 2.** Word frequency distribution for the job title.

Word distributions for full text are given in figure 3. Experience, work, and team are the most frequently used words in full text.

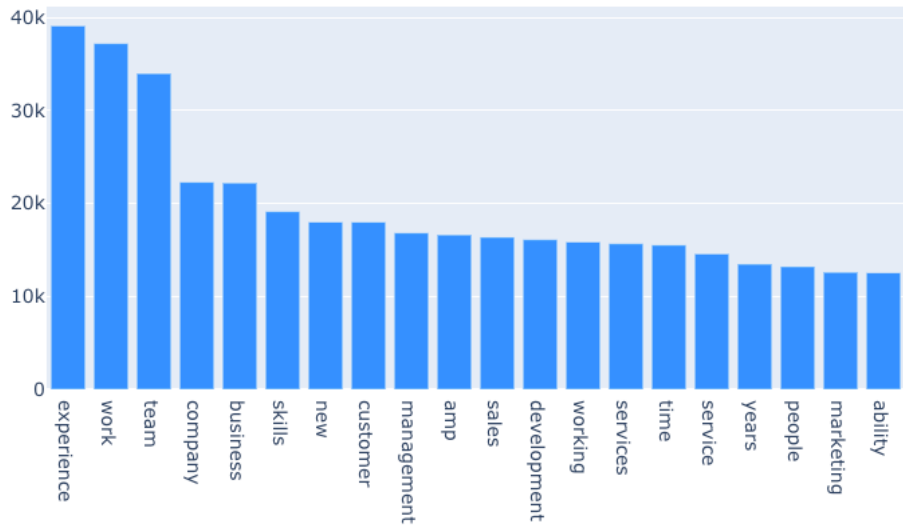

**Figure 3.** Word frequency distribution for full text.

### 2.4 Data processing

Data cleaning removes unnecessary characters, symbols, and HTML tags from the dataset. This includes the removal of punctuation marks, special characters, and any unnecessary white spaces.

Text preprocessing: We used the TextVectorization layer to transform the text data into a numerical format. This layer can be used to convert text data to a fixed-length vector representation, making it easier to feed into a machine-learning model.

Categorical Value handling: One-hot encoding is a common technique for handling categorical variables in machine learning models, including deep learning models like Bidirectional LSTMs. We used one-hot encoding to transform categorical variables such as job location, industry, employment type, and education into a numerical format that the model can understand.

### III. MODELING

Our modeling approach compares the performance of a bidirectional Long Short-Term Memory (BiLSTM) network with ensemble learning methods, including Random Forest, LightGBM, and Gradient Boosting Machine (GBM), for detecting fake job postings.

### 3.1 Ensemble Models

- Random Forest: An ensemble learning method that constructs multiple decision trees and aggregates their results. Each tree in the forest returns class prediction, and the class that gets the most votes become the model's prediction.
- LightGBM: A gradient boosting framework that uses tree-based learning algorithms optimized for higher efficiency and lower memory usage.
- Gradient Boosting Machine (GBM): A machine learning technique that builds an ensemble of weak prediction models, typically decision trees, and optimizes them through gradient descent. With gradient boosting, each predictor improves the previous predictor by reducing its error.

### 3.2 Bidirectional LSTM

We implement a Bidirectional LSTM-based model to process the textual information in job postings. The model architecture is depicted in the figure.

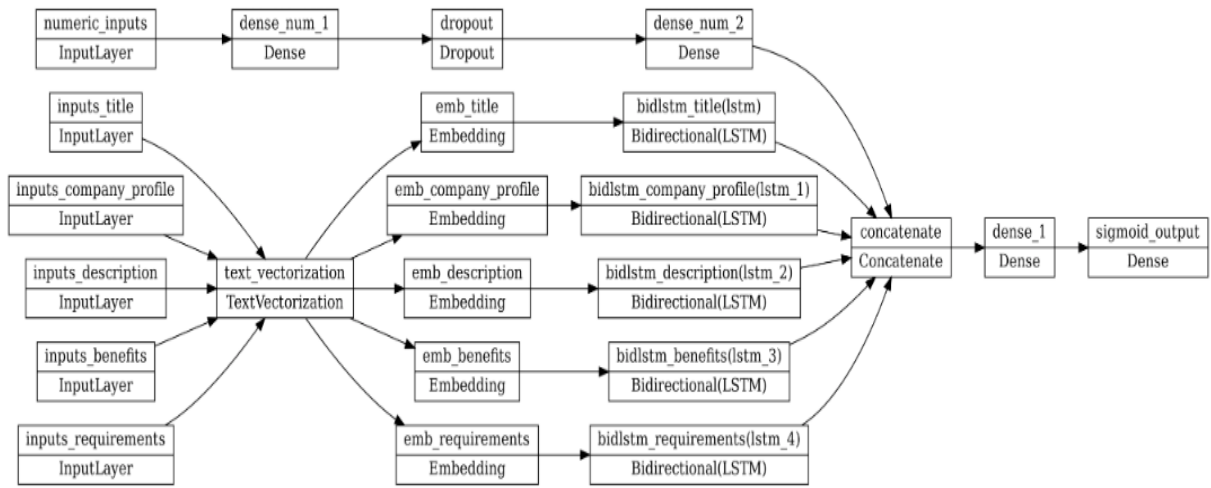

**Figure 4.** Bidirectional LSTM model architecture.

The model architecture consists of multiple layers, as given in figure 4.

1. Numerical Input Layer: This layer receives numerical features extracted from the job postings, such as salary range, job location, and company size.

2. Text Input Layer: This layer receives textual features, including job titles, company profiles, job description, benefits, and requirements.

3. Text Vectorization Layer: This layer converts the textual input into fixed-size numerical vectors. It tokenizes the text and maps the tokens to integers based on a pre-defined vocabulary.

4. Embedding Layer: This layer converts the integer-encoded tokens into dense vectors of fixed size. These dense vectors are continuous representations of words that capture their semantic meaning.

5. Bidirectional LSTM Layers: The bidirectional LSTM layers process the sequence of embedded word vectors in both forward and backward directions. This allows the model to capture context from both the past and the future, resulting in a richer understanding of the input text.

6. Merging Layer: This layer combines the output of the bidirectional LSTM layers with the numerical input features. The combined features are then used for further processing.

7. Dense Layer: A fully connected dense layer maps the combined features to a higher-level representation. This layer uses a ReLU (rectified linear unit) activation function to introduce non-linearity and improve the model's expressive power.

8. Sigmoid Activation Layer: The final layer in the model is a dense layer with a single neuron and a sigmoid activation function. This layer outputs a value between 0 and 1, representing the probability that a given job posting is fake. Based on this probability, a threshold can be set to classify postings as legitimate or fraudulent.

### 3.3 Model Training

We train the model using a labeled dataset of genuine and fake job postings. The dataset is split into training, validation, and testing sets to evaluate the model's performance. We randomly split the training data by 20% as validation and test data. The base models were trained with default hyperparameters. BiLSTM has been trained up to 25 epochs. Early stopping was achieved at six epochs. Loss and accuracy were used to determine early stop criteria.

### 3.4 Model Evaluation

To evaluate and compare the performance of our models, we use these metrics and interpret the results as follows:

- Accuracy: It measures how often the model correctly classified fake and real job postings. The higher the accuracy, the better the model's performance.
- ROC curve: It measures how well the model can distinguish between the positive and negative classes.
- Precision: It measures how many predicted job postings were fake. A high precision indicates that the model is good at identifying fake job postings but may miss some real ones.

- Recall: It measures how many fake job postings were correctly identified by the model. A high recall indicates that the model is good at identifying most fake job postings but may also have a high false positive rate.
- F1 score: It is a way to balance precision and recall, and a high F1 score indicates that the model is good at both identifying fake job postings and avoiding false positives.

By following this modeling approach, we develop a Bidirectional LSTM-based model capable of accurately identifying fraudulent job advertisements on online platforms. In the subsequent sections, we present our experiments' results and demonstrate our approach's efficacy compared to other state-of-the-art methods.

## IV.     RESULTS AND DISCUSSION

In this section, we present the results of our experiments, which demonstrate the effectiveness of our proposed Bidirectional LSTM model in detecting fake job postings. We evaluate the model's performance using a range of metrics, including accuracy, precision, recall, and F1-score and compare its performance to other state-of-the-art methods.

### 4.1 Bidirectional LSTM model performance

The bidirectional LSTM model achieved the following performance metrics on the test dataset, as given in Table 2.

**Table 2.** Bidirectional LSTM Performance Metrics.

| Model | auroc | accuracy | precision | Recall | f1 score |
|---|---|---|---|---|---|
| Bidirectional LSTM | 0.91 | 98.71 % | 0.89 | 0.83 | 0.86 |

We used a confusion matrix to measure the model's efficiency in predicting the labels. The confusion matrix is a table that shows the performance of a classification model by comparing the predicted class labels to the actual class labels. Out of the 3576 records in the test set, the BiLSTM model correctly identified labels for 3530 records. In addition, 144 true positives and 3386 true negatives were labeled correctly. The confusion matrix for the BiLSTM model is shown in Figure 5.

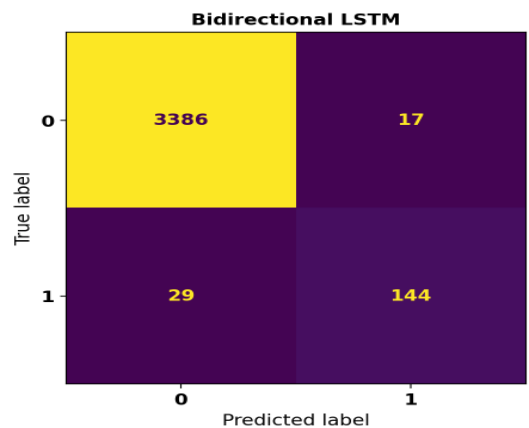

**Figure 5.** Bidirectional LSTM model confusion matrix.

### 4.2 Ensemble models performance

Ensemble models achieved the following performance metrics on the test dataset, as given in Table 3.

**Table 3.** Ensemble models Performance Metrics.

| Model | auroc | accuracy | precision | Recall | f1 score |
|---|---|---|---|---|---|
| Light GBM | 0.83 | 98.18% | 0.94 | 0.66 | 0.77 |
| Random Forest | 0.70 | 97.04% | 0.97 | 0.40 | 0.57 |
| Gradient Boosting | 0.68 | 96.67% | 0.86 | 0.37 | 0.62 |

Light GBM performed best among ensemble models achieving an accuracy of 98.18%. Out of the 3576 records in the test set, the Light GBM model correctly identified labels for 3511 records. In addition, 115 true positives and 3396 true negatives were labeled correctly. The confusion matrix for ensemble models is shown in Figure 6.

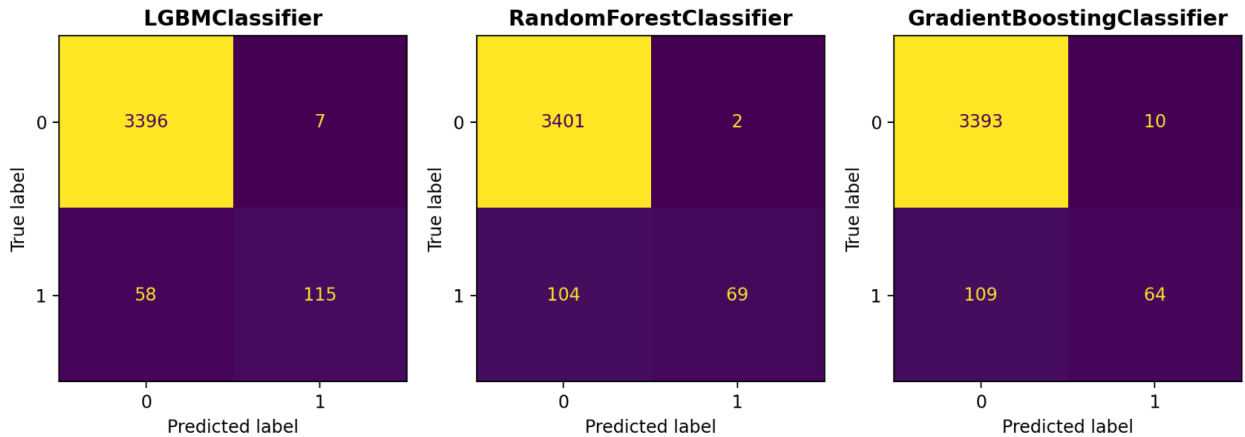

**Figure 6.** Ensemble models confusion matrix.

### 4.3 Discussion

Our experiments demonstrate the efficacy of our model in accurately identifying fraudulent job advertisements, outperforming other state-of-the-art methods. However, several challenges and areas for future research warrant further discussion.

**Class Imbalance:**

One of the main challenges in detecting fake job postings is the inherent class imbalance in the dataset, with a significantly lower number of fake job postings than genuine ones. This imbalance may lead to biased models that favor the majority class. While our Bi-LSTM model has demonstrated promising results, future work should explore techniques to address class imbalance more effectively, such as oversampling, undersampling, or employing cost-sensitive learning approaches.

**Transfer Learning and Domain Adaptation:**

Our proposed model is trained and tested on a specific dataset of job postings. However, the model's performance on job postings from other sources or in different languages is uncertain. Exploring transfer learning and domain adaptation techniques could help improve the model's generalization capabilities, allowing it to perform well across various platforms and languages.

**Interpretability and Explainability:**

Deep learning models, including Bi-LSTM networks, are often criticized for their lack of interpretability and explainability. However, understanding the rationale behind a model's decision is crucial, especially when dealing with sensitive information such as job postings. Therefore, future work should incorporate interpretability and explainability into the model, ensuring that users can understand and trust the model's decisions.

**Ethical Considerations:**

The deployment of automated fake job posting detection systems raises ethical concerns, such as potential biases in the training data or the risk of false positives and negatives. Ensuring that these models are transparent, fair, and robust is crucial to minimize the risk of unintended consequences. Future research should address these ethical concerns and develop guidelines for the responsible use of machine learning techniques in detecting fake job postings.

## V.    CONCLUSION

In this article, we have presented a novel approach to detecting fake job postings using Bidirectional LSTM networks. Through a series of experiments, we demonstrated that our proposed model could effectively identify fraudulent job advertisements on online platforms, achieving high precision and recall compared to other state-of-the-art methods. The key advantages of our approach lie in the ability of the Bi-LSTM model to capture complex patterns and contextual information within the textual data of job postings. By leveraging these capabilities, our model offers a more accurate and reliable solution to the growing issue of fake job postings, ultimately providing users with a safer and more trustworthy job-seeking experience.

Our work contributes to the ongoing efforts to combat online job fraud and highlights the potential of Bidirectional LSTM networks in addressing deceptive content detection problems. In addition, we hope our findings inspire further research and development in this area, ultimately leading to more effective solutions for ensuring the integrity of online job portals.

## VI. CONFLICTS OF INTEREST

The author declares no conflicts of interest regarding the publication of this paper.

**Code Availability**

The code used for this study is available for reference:

https://github.com/aravindsp/fake-job-postings-detection/blob/main/code/fake-job-postings-detection-bidlstm.ipynb